\pdfoutput=1

\documentclass[11pt]{article}

\usepackage{acl}
\usepackage{xcolor}

\usepackage{times}
\newcommand{\Rom}[1]{\uppercase\expandafter{\romannumeral #1\relax}}
\newcommand{\rom}[1]{\lowercase\expandafter{\romannumeral #1\relax}}

\usepackage[T1]{fontenc}

\usepackage[utf8]{inputenc}
\usepackage{graphicx}
\graphicspath{ {./Figures/} }
\usepackage{lipsum}
\usepackage{booktabs}
\usepackage{graphicx}
\usepackage{multirow}
\usepackage{microtype}

\title{When do you need Chain-of-Thought Prompting for ChatGPT? }

\author{Jiuhai Chen* \\
  University of Maryland \\
  \texttt{jchen169@umd.edu} \\\And
  Lichang Chen*  \\
  University of Maryland \\
  \texttt{bobchen@umd.edu} \\ \And
  Heng Huang \\
  University of Maryland \\
  \texttt{heng@umd.edu} \\ \And
  Tianyi Zhou \\
  University of Maryland \\
  \texttt{tianyi@umd.edu}
  }

\begin{document}
\maketitle
\def\thefootnote{\textbf{*}}\footnotetext{Equal Contribution.}\def\thefootnote{\arabic{footnote}}
\begin{abstract}

Chain-of-Thought~(CoT)~prompting can~effectively elicit complex multi-step reasoning from Large Language Models~(LLMs). For example, by simply adding CoT instruction ``Let's think step-by-step'' to each input query of MultiArith dataset, GPT-3's accuracy can be improved from 17.7\% to 78.7\%. 
However, it is not clear whether CoT is still effective on more recent instruction finetuned (IFT) LLMs such as ChatGPT. 
Surprisingly, on ChatGPT, CoT is no longer effective for certain tasks such as arithmetic reasoning while still keeping effective on other reasoning tasks. Moreover, on the former tasks, ChatGPT usually achieves the best performance and spontaneously generates CoT even without being instructed to do so.
Hence, it is plausible that ChatGPT has already been trained on these tasks with CoT and thus memorized the instruction so it implicitly follows such an instruction when applied to the same queries, even without CoT. 
Our analysis reflects a potential risk of overfitting/bias toward instructions introduced in IFT, which becomes more common in training LLMs.
In addition, it indicates possible leakage of the pretraining recipe, e.g., one can verify whether a dataset and instruction were used in training ChatGPT. 
Our experiments report new baseline results of ChatGPT on a variety of reasoning tasks and shed novel insights into LLM's profiling, instruction memorization, and pretraining dataset leakage.\looseness-1
\end{abstract}

\section{Introduction}


\begin{figure}[htp]
\includegraphics[width=7.5cm]{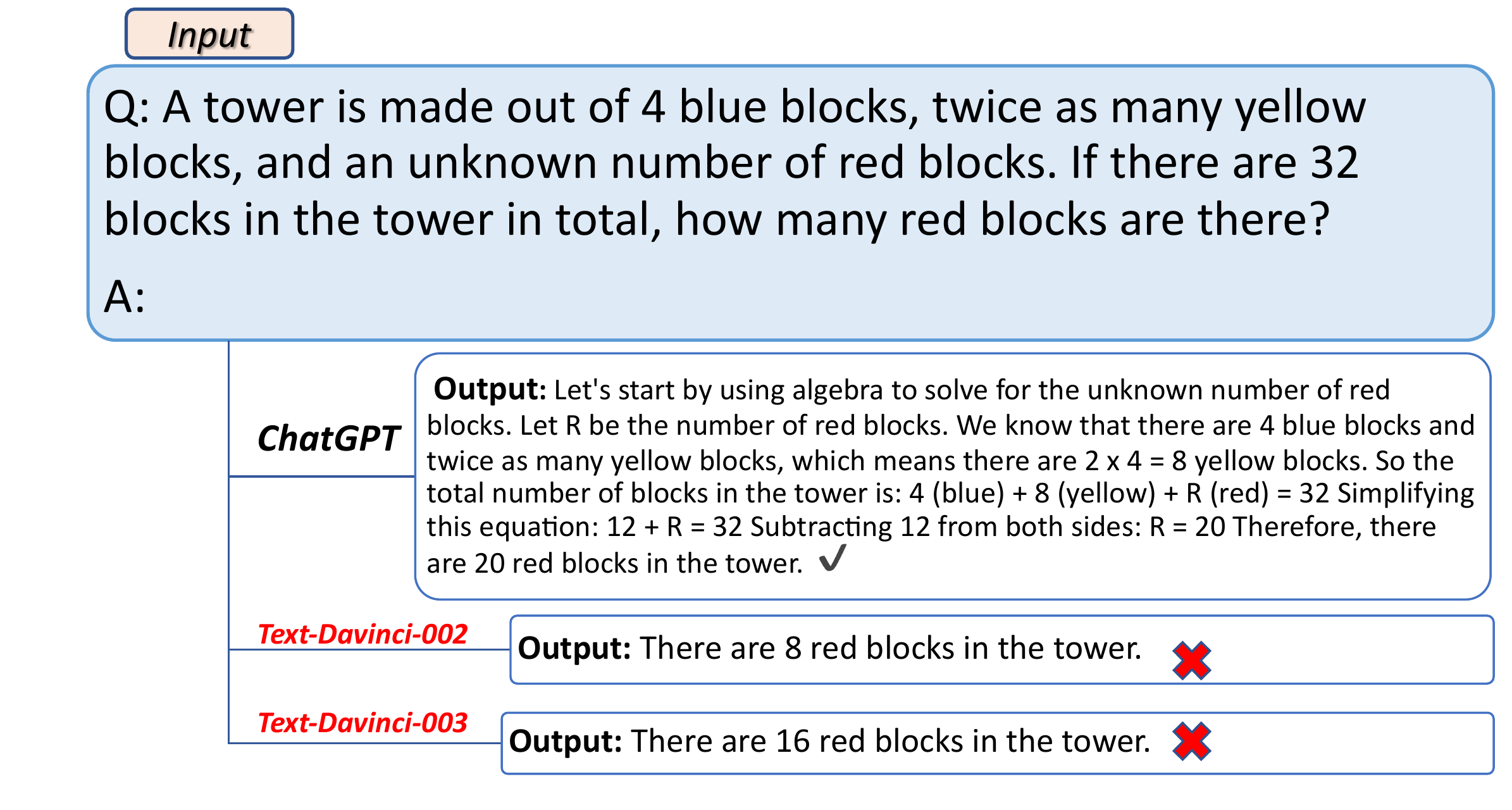}
\centering
\caption{\label{fig:model_output_math} An example of arithmetic reasoning by different LLMs when \textbf{prompting without instruction}, i.e., the input only contains the original question. While Text-Davinci-002 and  Text-Davinci-003 generate wrong answers, ChatGPT spontaneously generates a sequence of CoT reasoning steps leading to a correct answer. 
}
\vspace{-1em}
\end{figure}

\par Zero-shot generalization of large language models (LLMs) can be significantly improved by letting LLMs follow instructions for different tasks. For example, a widely-used instruction for reasoning tasks is chain-of-thoughts (CoT) prompting~\cite{kojima2022large}, which can improve GPT-3's math reasoning accuracy from 17.7\% to 78.7\% (on MultiArith dataset~\cite{roy2016multiarith}) by simply appending the instruction ``Let's think step-by-step'' to each query. 
In order to reinforce this capability, more recent LLMs such as~InstructGPT~\cite{InstructGPT} are trained using instruction finetuning (IFT) on thousands of tasks associated with different instructions. 
ChatGPT~\cite{van2023chatgpt, jiao2023chatgpt}, a state-of-the-art conversational agent designed to engage in erudite discourse and provide profound insights across a myriad of subjects, was trained using IFT and reinforcement learning with human feedback~(RLHF)~\cite{christiano2017deep}. 
Considering these changes made to the training strategies, a natural inquiry then arises: is CoT (or other instructions) still effective on ChatGPT (or other LLMs trained by IFT and RLHF)? Since OpenAI only released the API for ChatGPT very recently, it is challenging to study this problem without extensive evaluation of ChatGPT on multiple tasks or access to ChatGPT's weights/training data.\looseness-1


\par
\begin{figure*}[h]
\centering

\begin{minipage}{1.0\textwidth}
\centering
\includegraphics[width=0.9\linewidth]{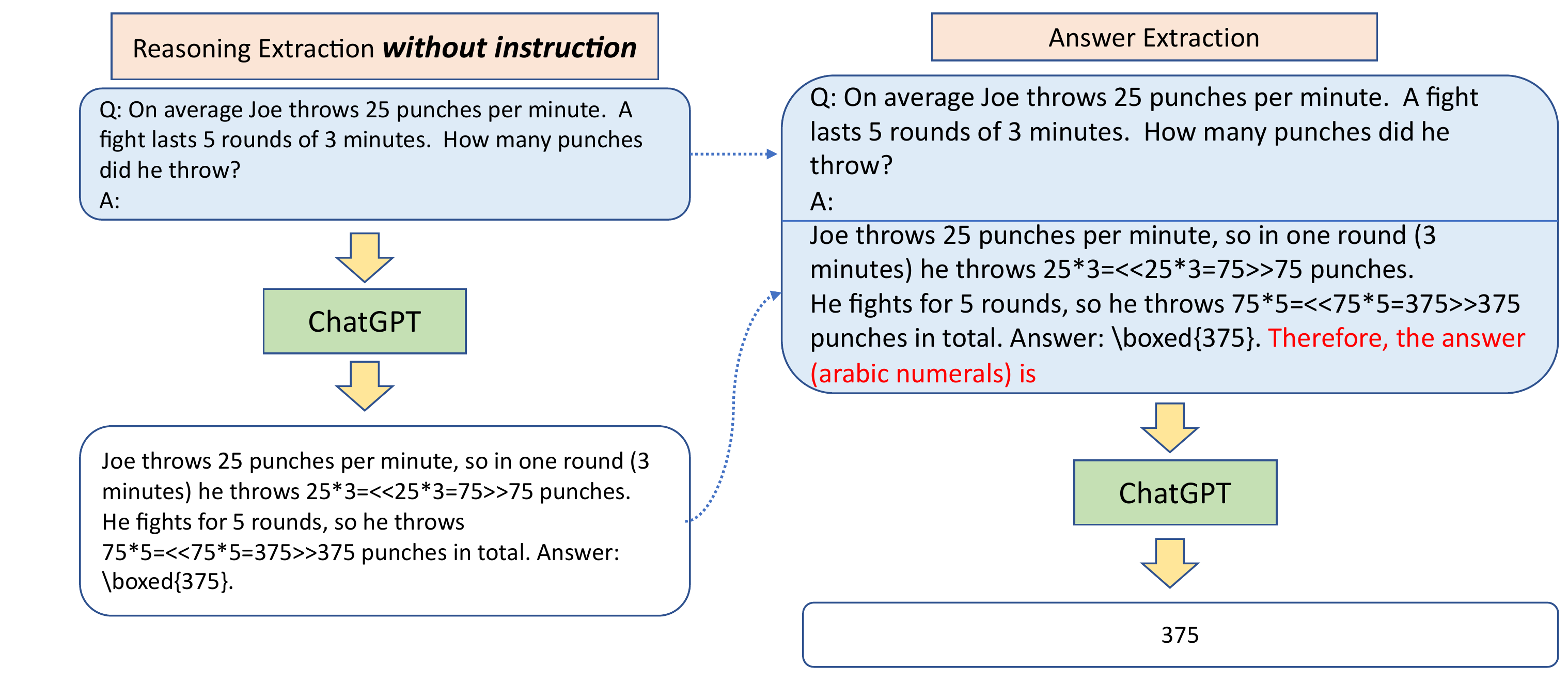}
\caption{Zero-shot reasoning without instruction (the first query) followed by prompting with trigger words.}
\label{fig:zero_shot_wo_prompt}
\end{minipage}
\bigskip{}\bigskip{}
\begin{minipage}{1.0\textwidth}
\centering
\includegraphics[width=0.9\linewidth]{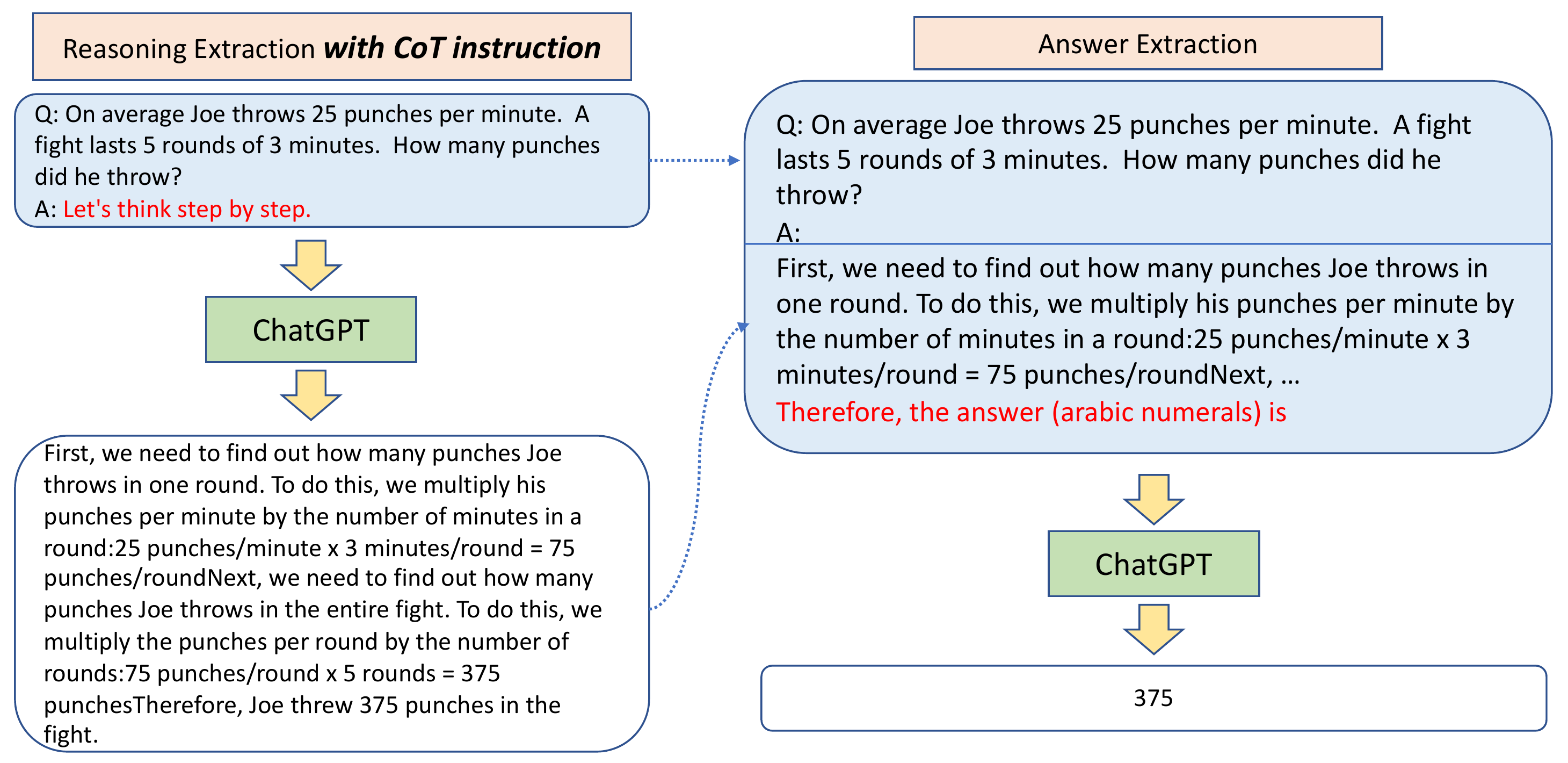}
\caption{Zero-shot reasoning with CoT instruction (the first query) followed by prompting with trigger words. }\label{fig:zero_shot_cot}
\end{minipage}
\end{figure*}

 To study the aforementioned problem, we conduct experiments on a variety of reasoning benchmarks widely used in previous literature by comparing three zero-shot learning strategies on GPT-3 and ChatGPT: (1) \textbf{Zero-shot with trigger words only}: the input is a question followed by a prompt of trigger words; (2) \textbf{Zero-shot without instruction}: LLM is only given a question in the first stage without any instruction and its output is then included in the second stage, which applies (1) to produce the final answer (see Figure \ref{fig:zero_shot_wo_prompt}); and (3) \textbf{Zero-shot with CoT instruction}: the same two-stage strategy as (2) except adding a CoT instruction (i.e., ``Let's think step-by-step''~\citet{kojima2022large}) after the question in the first stage (see Figure \ref{fig:zero_shot_cot}). On previous LLMs such as GPT-3, (3) significantly improves a variety of reasoning tasks' performance without using any training data. 

\begin{figure}[ht]
\includegraphics[width=0.49\textwidth]{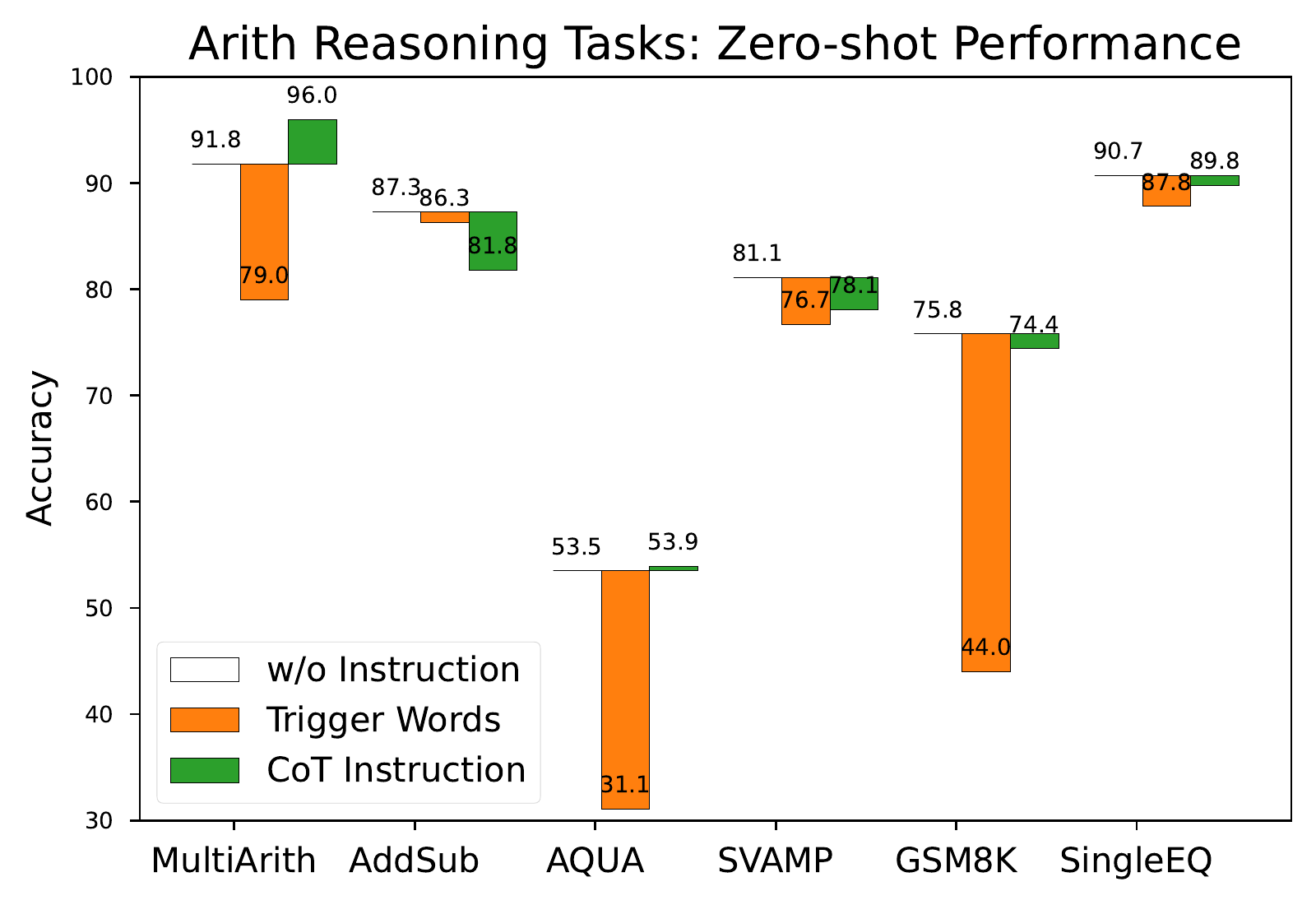}
\centering
\vspace{-1em}
\caption{\label{fig:results_arithmetic} 
Comparison of the three prompting strategies in Section~\ref{sec:prompts} when applied to ChatGPT for \textbf{six arithmetic reasoning tasks}.
 }
 \vspace{-1em}
\end{figure}

\begin{figure}[ht]
\includegraphics[width=0.49\textwidth]{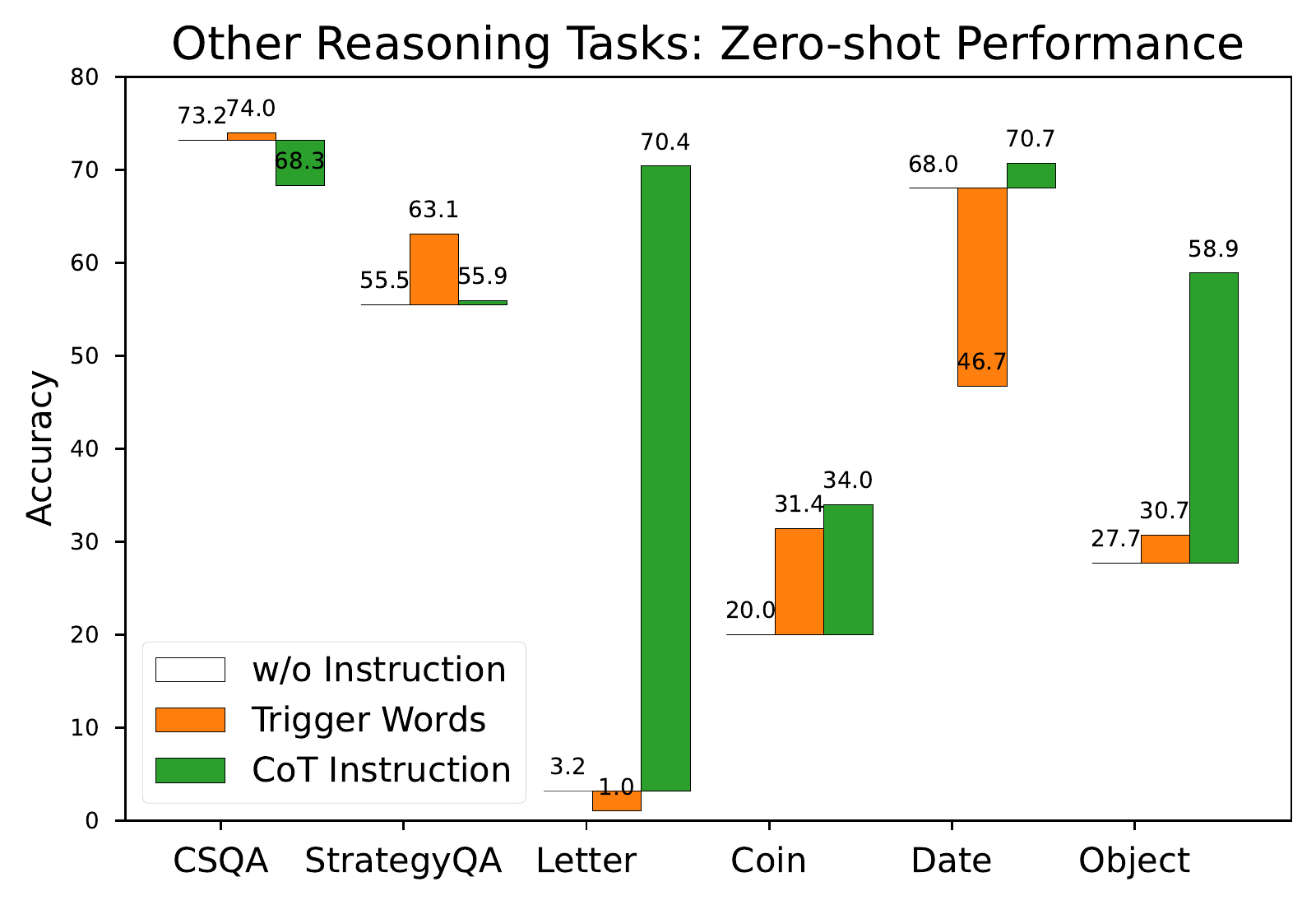}
\centering
\vspace{-1em}
\caption{\label{fig:results_other} 
Comparison of the three prompting strategies in Section~\ref{sec:prompts} when applied to ChatGPT for \textbf{common sense, symbolic, and other two reasoning tasks}.
 }
\end{figure}

\par\textbf{Our Observations.}~Surprisingly, the comparison of the three zero-shot strategies on ChatGPT exhibits very different patterns from that on GPT-3. Specifically, on some datasets and reasoning tasks, ChatGPT without any instruction given (i.e., (2) Zero-shot without instruction) can spontaneously produce CoT steps for most questions and produce high-quality answers, while CoT instruction (i.e., (3) Zero-shot with CoT) cannot further improve the accuracy or even cause degradation. In contrast, CoT instruction always brings significant improvement to GPT-3 on all the evaluated reasoning tasks. 

\par\textbf{Our Insights.}~Considering that ChatGPT was trained using IFT, we posit that the observed difference on ChatGPT is caused by its memorization of CoT instruction during IFT, which enforces it to follow an implicit instruction when encountering the same (or similar) question, even without using any instruction. This implies a risk of dataset and instruction leakage, i.e., we might be able to verify if a dataset and instruction were utilized in an LLM's pretraining recipe by only querying the LLM as a black box. Thereby, we take the first step towards dataset inference attack~(DIA) for LLMs (section~\ref{sec: mia & dia}). DIA can play an important role in language model profiling, which is an analytical approach employed to examine and characterize the linguistic capabilities, strengths, and limitations of LLMs. Accurate profiles of LLMs can significantly facilitate the selection of LLMs for specified tasks in practical applications without costly evaluations.

\par In addition, for the first time, our empirical analysis provides an extensive and precise evaluation of ChatGPT's zero-shot learning capabilities on a variety of representative reasoning tasks when using different prompting strategies, which set up new baseline results for future works and improve the understanding of ChatGPT on reasoning tasks. 

The main takeaways can be summarized by:
\begin{enumerate}
\itemsep0em 
    \item ChatGPT spontaneously generates intermediate steps for arithmetic reasoning tasks even without instruction (see Figure \ref{fig:model_output_math}).
    \item Unlike GPT-3 and previous LLMs, CoT instruction is not helpful or even detrimental to ChatGPT on arithmetic reasoning tasks.
    \item On other reasoning tasks (except arithmetic ones), ChatGPT does not exhibit the above patterns and behaves similarly to GPT-3.  
    \item It is plausible that ChatGPT utilized arithmetic reasoning with CoT instruction for IFT and memorizes the instruction and data.
    \item Our observations imply a leakage of the pretraining recipe, overfitting to instructions of IFT, and a dataset inference attack. 
\end{enumerate}


\section{Related Work}
\subsection{ChatGPT} 
ChatGPT, a cutting-edge conversational AI, has been widely acknowledged for its groundbreaking improvement to the AI generate-context~(AIGC)\footnote{\url{https://www.forbes.com/sites/forbestechcouncil/2023/03/09/will-chatgpt-solve-all-our-problems/?sh=6f3e25494833}} and also suggests a new era of AI research is coming~\cite{jiao2023chatgpt, van2023chatgpt}. It not only shows exceptional cognitive abilities~(In standardized tests like SAT, GRE\footnote{\url{https://twitter.com/emollick/status/1635700173946105856}}, it could even obtain better scores than we human beings. ) but also human-level context writing skill: even with watermarks for LLM, it is hard to be reliably distinguished from the context created by human beings~\cite{sadasivan2023can, krishna2023paraphrasing}.

\subsection{Chain-of-Thought Prompting}
Chain-of-thought prompting~(CoT)~\cite{wei2022chain, kojima2022large} is a two-tiered querying strategy applied to LLMs for zero-shot and few-shot reasoning. CoT prompts (an instruction or a few CoT exemplars) can elicit a sequence of intermediate reasoning steps for each query from the LLMs. Built upon~\citet{wei2022chain}, numerous investigations have enhanced the standard CoT through various strategies, including least-to-most prompting~\cite{zhou2022least}, self-consistency~\cite{wang2022self}, bootstrapping~\cite{boostrap}, selecting better demos~\cite{chen2023takes, verifier}, etc. These advancements have significantly bolstered the performance of CoT prompting in tackling intricate tasks. We will focus on zero-shot CoT (i.e., instruction prompting) in this paper and study whether it is still effective on ChatGPT that might already pre-trained with CoT instruction via IFT. 


\subsection{Membership/Dataset Inference Attack}
\label{sec: mia & dia}
Membership inference (MI) attack has been studied for image classification models~\cite{mia, emia}. It aims to determine whether a data record was included in the model's training dataset or not, with only black-box access to the model is allowed. There are two main categories of MI: confidence vector~\cite{HayesMDC19MIA, Salem0HBF019MIA}, where the attacker has access to the model's predicted confidence score, and label-only~\cite{label-only}. 
However, inference attack for LLMs is more challenging because (1) 
the training corpus for LLMs is much larger, e.g., 499 billion tokens for GPT-3\footnote{\url{https://lambdalabs.com/blog/demystifying-gpt-3}} compared to 14 million images for ImageNet~\cite{deng2009imagenet}; and (2) the LLM output is combinatorial and highly structured while the output space of text/image classification is finite and much simpler. Hence, MI of a single data record for LLMs would be like finding a needle in a haystack and we instead investigate the leakage of training datasets.
\paragraph{Dataset Inference.} For an LM~$\mathcal{M}$ trained on a large training corpus via pretraining or IFT, i.e.,
\begin{equation}
    \mathcal{C} = \{D_1, D_2, \ldots, D_N \} \cup \{I_1, I_2, \ldots, I_N\},
\end{equation}
where $D_i$ and $I_i$ represent the dataset and instruction, respectively.
Given a dataset~$D_i$, dataset inference aims to verify if $D_i \in \mathcal{C}$ or $I_i \in \mathcal{C}$.

\section{Zero-shot Reasoning on ChatGPT}

\subsection{Prompting Strategies}\label{sec:prompts}
Recent LLMs exhibit great potential in performing zero-shot learning~\cite{zsl-survey, xian2017zero} without any training any model parameters via prompting. We compare three zero-shot reasoning strategies applied to ChatGPT and GPT-3 on a variety of reasoning tasks.
\paragraph{(1) Zero-shot with Trigger Words.} We query the LLM $M$ by adding trigger words $T$ after each input question $Q$, i.e., $A=M([Q;T])$. We follow a list of trigger-word prompts from \citet{kojima2022large}, e.g, ``the answer (Arabic numerals) is'' for arithmetic reasoning tasks. 
\paragraph{(2) Zero-shot without instruction.} We query the LLM twice where the first query is composed of the original question only (without any instruction), $C=M([Q])$, and its output $C$ is included in the second query's input and appended with trigger words to produce the final answer, i.e., $A=M([C;Q;T])$ (see Figure \ref{fig:zero_shot_wo_prompt}). 

\paragraph{(3) Zero-shot with CoT instruction.} We query the LLM twice as the strategy above except by adding a CoT instruction $P$ (i.e., ``Let's think step-by-step'') to the original question in the first query. Hence, the first query is $C=M([Q;P])$ and the second is $A=M([C;Q;T])$ (see Figure \ref{fig:zero_shot_cot}). 

\subsection{Tasks \& Datasets}
We conduct a series of experiments on a variety of reasoning benchmarks: 
\textbf{Arithmetic Reasoning:} GSM8K \cite{cobbe2021GSM8K}, MultiArith \cite{roy2016multiarith}, AddSub \cite{hosseini2014AddSub}, SVAMP \cite{patel2021nlp}, AQuA \cite{ling2017program} and SingleOp \cite{wei2022chain}. 
\textbf{Commonsense Reasoning:} CSQA \cite{talmor2018commonsenseqa} and StrategyQA \cite{geva2021did}. \textbf{Symbolic Reasoning:} Last Letter Concatenation (Last Letter) \cite{wei2022chain} and Coin-flip \cite{wei2022chain}. Other reasoning tasks: Date understanding (Date) and Tracking Shuffled Objects (Object) \cite{wei2022chain}. The overall statistics are listed in Table \ref{tab:data}.

\begin{table}[ht]
\centering
\scalebox{0.85}{
\begin{tabular}{lll}
\toprule
 Dataset & Task &  \# Query  \\
 \midrule
GSM8K & Arithmetic  & 1319   \\
MultiArith   & Arithmetic  & 600   \\
AddSub & Arithmetic  &  395  \\
SVAMP  & Arithmetic  & 1000  \\
AQuA & Arithmetic &  254 \\
SingleOp  & Arithmetic  & 508  \\
CSQA  & Commonsense &  1221  \\
StrategyQA & Commonsense &  2290  \\
Last Letter & Symbolic  & 500  \\
Coin-flip  & Symbolic  & 500  \\
Date & Other  & 369  \\
Object & Other  & 750  \\
\bottomrule
\end{tabular}}
\caption{\label{tab:data}
Statistics of tasks and datasets. \# Query is the number of test samples.}
\end{table}


\begin{table*}[htbp]
\centering
\scalebox{0.9}{%
\begin{tabular}{|l|l|cccccc|}
\hline
 \multirow{2}{*} {LLM} & \multirow{2}{*} {Prompt} & \multicolumn{6}{|c|}{\textbf{Arithmetic}}  \\
  &  & MultiArith & AddSub  & AQUA-RAT & SVAMP & GSM8K & SingleEQ \\
    \hline
\multirow{2}{*}{GPT-3} 
& Trigger words & 17.7 & 72.2 & 22.4 & 58.8 & 10.4 & 74.6 \\
& CoT instruction & 78.7 & 69.6 & 33.5 & 62.1 & 40.7 & 78.0 \\ 
  \hline
\multirow{3}{*}{ChatGPT}
& Trigger words & 79.0 & 86.3 & 31.1 & 76.7 & 44.0 & 87.8 \\
& No instruction & 91.8 & 87.3 & 53.5 & 81.1 & 75.8 & 90.7 \\

& CoT instruction & 96.0 & 81.8 & 53.9 & 78.1 & 74.4 & 89.8 \\ 
\hline
\hline
\end{tabular}}
\caption{Zero-shot reasoning accuracy (\%) of GPT-3 and ChatGPT on \textbf{six arithmetic reasoning tasks} when using different prompting strategies in Section~\ref{sec:prompts}.}
\label{tab:mathqa}
\end{table*}

\begin{table*}[htbp]
\centering
\scalebox{0.9}{
\begin{tabular}{|l|l|cc|cc|cc|}
\hline
 \multirow{2}{*} {LLM} & \multirow{2}{*} {Prompt} & \multicolumn{2}{|c|}{\textbf{Common Sense}} & \multicolumn{2}{|c|}{\textbf{Symbolic Reasoning}} & \multicolumn{2}{|c|}{\textbf{Other Reasoning}} \\
  &  & CSQA & StrategyQA  & Letter & Coin & Date & Object \\
\hline
  \multirow{2}{*}{GPT-3} 
  & Trigger words & 68.8 & 12.7  & 0.2 & 12.8 & 49.3 & 31.3 \\
& CoT instruction & 64.6 & 54.8  & 57.6 & 91.4 & 67.5 & 52.4   \\
  \hline
  \multirow{3}{*}{ChatGPT} 
& Trigger words & 74.0 & 63.1  & 1.0 & 31.4 & 46.1 & 30.7   \\
& No instruction & 73.2 & 55.5  & 2.6 & 20.0 & 68.0 & 27.7 \\
& CoT instruction & 68.3 & 55.9  & 70.4 & 34.0 & 70.7 & 58.9  \\ 
\hline
\hline
\end{tabular}}
\caption{
Zero-shot reasoning accuracy (\%) of GPT-3 and ChatGPT on \textbf{common sense, symbolic, and other two reasoning tasks} when using different prompting strategies in Section~\ref{sec:prompts}.
}
\label{tab:csqa}
\end{table*}

\section{Discoveries and Analysis}
\subsection{ChatGPT vs. GPT-3 with different prompting strategies and tasks}\label{sec:observations}

\begin{figure}[t]
\includegraphics[width=7.5cm]{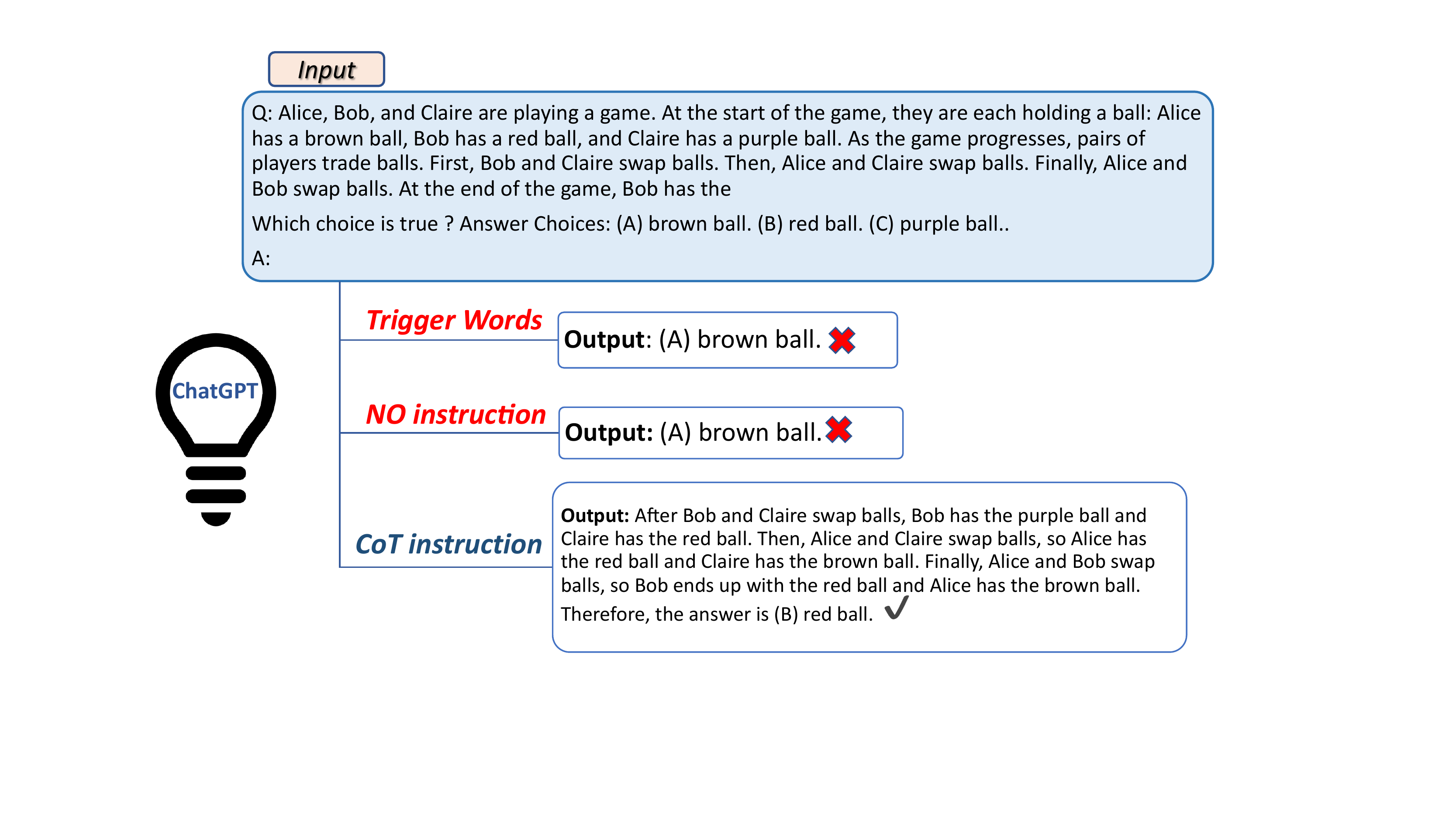}
\centering
\caption{\label{fig:model_output_object} Comparison of the three prompting strategies in Section~\ref{sec:prompts} applied to \textbf{ChatGPT on an example of Tracking Shuffled Object reasoning}. Unlike arithmetic reasoning, ChatGPT cannot spontaneously generate step-by-step rationale without instruction. Meanwhile, CoT prompting elicits the step-by-step reasoning capability of ChatGPT and results in a correct answer. \looseness-1
}
\end{figure}

We compare the three prompting strategies for zero-shot reasoning introduced in Section~\ref{sec:prompts} when applied to GPT-3 and ChatGPT. The results are reported in Table~\ref{tab:mathqa}-\ref{tab:csqa}. Interestingly, we have different observations on the two LLMs and they exhibit different patterns on different reasoning tasks:
\begin{itemize}
\itemsep0em
    \item On almost all the reasoning tasks in Table~\ref{tab:mathqa}-\ref{tab:csqa} (except CSQA), CoT instruction prompt consistently improves the zero-shot reasoning accuracy of GPT-3. This is consistent with previous work's observations~\citet{kojima2022large}. 
    \item However, when applied to ChatGPT, it performs even worse than prompting with ``no instruction'' on four out of the six tasks and only brings $+0.4\%$ improvement on one task (i.e., AQUA-RAT). Surprisingly, ChatGPT still generates CoT reasoning steps in the first prompting even with ``no instruction'' given, as shown in Figure~\ref{fig:model_output_math}. In other words, ChatGPT spontaneously performs better CoT reasoning when not explicitly instructed to ``think it step-by-step''. This is contrary to GPT-3. 
    \item On most of the non-arithmetic reasoning tasks in Table~\ref{tab:csqa} (except CSQA), as shown in Figure~\ref{fig:model_output_object}, CoT instruction significantly improves ChatGPT's zero-shot reasoning accuracy on five out of the six tasks. This is consistent with GPT-3 and other previous LLMs.
    \item On the two commonsense reasoning tasks, single-step prompting with only trigger words achieves the best performance with ChatGPT. It also performs the best on CSQA with GPT-3. \looseness-1
\end{itemize}
Therefore, unlike the consistent improvement by CoT on GPT-3, whether CoT instruction can improve ChatGPT or not varies across datasets and depends on at least the style of datasets (e.g., arithmetic or not). This leads to a fundamental key question: \textbf{when do you need CoT prompting for ChatGPT, and why?}

\subsection{Memorization of Instructions}

A major difference between ChatGPT and previous LLMs, as shown in Section~\ref{sec:observations}, is that ChatGPT does not need CoT instruction (or any instruction) but can spontaneously produce step-by-step reasoning that leads to higher-quality answers for arithmetic reasoning. Moreover, CoT instruction can even deteriorate ChatGPT's arithmetic reasoning accuracy. On commonsense reasoning tasks, 
adding CoT instruction to the first query cannot noticeably improve ChatGPT's accuracy (on StrategyQA) or even results in degradation (CSQA). In fact, removing the entire first query and only applying the second one, i.e., the trigger-words prompting, results in the best accuracy of ChatGPT on these two tasks.  
However, such differences disappear on other reasoning tasks, and ChatGPT, similar to other LLMs such as GPT-3 and PaLM~\cite{kojima2022large}, tends to benefit a lot from CoT prompting and produces precise intermediate reasoning steps when given the CoT instruction. 


Hence, \textbf{on specific types of tasks, ChatGPT tends to follow an implicit CoT instruction even not being instructed to do anything}. Moreover, such implicit instruction-following performs even better than prompting with an explicit CoT instruction, which might be a distraction to the implicit one. However, this behavior does not happen on GPT-3 and we posit that the behavior is a consequence of the training strategy, more specifically, IFT for ChatGPT, because the main difference between ChatGPT and GPT-3 is that the former was trained on specified (task, instruction) pairs. In particular, IFT for ChatGPT might include most arithmetic reasoning tasks and CSQA for commonsense reasoning trained with the CoT instruction. During IFT, it is plausible that ChatGPT memorizes the CoT instruction and builds an inherent mapping from these types of reasoning questions to the memorized CoT instruction. Hence, when applied to the same tasks/datsets, even without any explicit instruction, ChatGPT still follows the memorized instruction associated with the tasks/datsets. 

The plausible instruction memorization of ChatGPT indicates a risk of overfitting or bias to instructions covered by IFT. Although IFT has been demonstrated to be effective on training LLMs towards more professional task solvers, the instructions and tasks covered by IFT are finite and possibly limited. Hence, it is unclear whether and how IFT trained LLMs generalize to new instructions and tasks. Moreover, a general-purpose LLM is expected to follow different instructions even for the same question. However, based on our observations on ChatGPT, it may have difficulty to follow other instructions if the assigned tasks are associated with certain specified instructions during IFT. 


\subsection{Pretraining Recipe Leakage via Dataset Inference Attack}

The analysis of instruction memorization above naturally implies a risk of pretraining recipe leakage and enables a dataset inference attack introduced in Section~\ref{sec: mia & dia}. Although the pretraining recipe, e.g., the tasks/datasets and their associated instructions in IFT, is not released for ChatGPT and more recent API LLMs, \textbf{we might be able to verify whether a (task, instruction) pair was used in their IFT or not} by looking for the difference between the prompting strategies in Section~\ref{sec:prompts}. 

For example, based on our observations, it is reasonable to posit that ChatGPT has been trained on arithmetic reasoning tasks and CSQA with the CoT instruction during IFT but not on the other reasoning tasks. Moreover, by comparing the difference between ``trigger words'' and ``CoT instruction'' in Table~\ref{tab:mathqa}-\ref{tab:csqa}, CoT prompting only brings significant improvement to a few arithmetic reasoning datasets (i.e., MultiArith, AQUA-RAT, GSM8K), the two symbolic reasoning datasets, and the other two reasoning datasets. This may indicate a dataset leakage, i.e., CoT prompting might not be able to further improve ChatGPT on the training data.  


Notably,  GPT-4 technical report \footnote{\url{https://arxiv.org/pdf/2303.08774.pdf}} confirmed that GPT-4 mixed in data from the training set of MATH and GSM8K to improve GPT-4's capability of mathematical reasoning, and they did used CoT prompting. Provided that GPT-4 surpasses ChatGPT on its advanced reasoning capabilities, it is highly plausible that ChatGPT was also trained on a certain amount of MATH datasets in the pre-training stage. That being said, more extensive analyses and rigorous tests are needed in the future to fully confirm these conjectures. 

\section{Conclusion}
 We investigate the reasoning capability of ChatGPT using three different prompting strategies and observe disparate behaviors of ChatGPT compared to previous LLMs such as GPT-3. Moreover, we found that such disparities highly depend on the task and prompt type. Specifically, different from GPT-3 whose zero-shot reasoning accuracy can be almost always improved by following a CoT instruction, ChatGPT without any instruction given in the prompt performs the best or at least comparably and surprisingly generates CoT reasoning steps for input questions spontaneously on most arithmetic and commonsense reasoning tasks. On the other hand, ChatGPT shows consistent patterns as GPT-3 and other LLMs on other reasoning tasks and benchmarks, on which CoT prompting usually brings significant improvement. 

 Given that a featured difference of ChatGPT comparing to GPT-3 and previous LLMs is IFT, i.e., finetuning on a set of (task, instruction) pairs, we posit that the isparities and spontaneous CoT behavior of ChatGPT observed on certain tasks are due to a memorization of the CoT instruction when training on these tasks during the IFT phase. In addition, our observations, upon further verification in the future, also indicate a risk of pretraining (IFT) recipe leakage and vulnerability to dataset inference attacks for the current API LLMs. Furthermore, our analyses underscore a number of fundamental challenges that need to be addressed in the future, e.g., whether and how the instruction-following capability of LLMs obtained through IFT can generalize to new tasks and instructions, how to choose the prompting strategies for different types of tasks, etc.
 
 
\bibliography{main}
\bibliographystyle{acl_natbib}



\end{document}